\documentclass{article}

\usepackage{arxiv}

\usepackage[utf8]{inputenc} 
\usepackage[T1]{fontenc}    
\usepackage{hyperref}       
\usepackage{url}            
\usepackage{booktabs}       
\usepackage{amsfonts}       
\usepackage{nicefrac}       
\usepackage{microtype}      
\usepackage{lipsum}

\usepackage{cite}
\usepackage{amsmath,amssymb,amsfonts}
\usepackage{graphicx}
\usepackage{textcomp}
\usepackage{hyperref}
\usepackage{slashbox}
\usepackage{booktabs}
\usepackage{multirow}
\usepackage{amsmath}
\usepackage{balance}
\usepackage{float}
\usepackage{algorithm}
\usepackage{algpseudocode}
\usepackage{caption}
\usepackage{subcaption}
\usepackage{url}
\usepackage{xcolor}
\usepackage{tikz}
\usepackage{forest}
\hypersetup{breaklinks=true}

\title{MaskedFace-Net - A dataset of correctly/incorrectly masked face images in the context of COVID-19}

\author{
  Adnane Cabani$^{1}$\thanks{\textbf{Adnane Cabani and Karim Hammoudi contributed equally to this work (corresponding authors).}} \\
  $^{1}$Normandie Univ, UNIROUEN\\
	ESIGELEC, IRSEEM\\
  76000 Rouen, France\\
  \texttt{adnane.cabani@esigelec.fr} \\	
   \And
Karim Hammoudi$^{2,3,*}$\\
$^{2}$Universit\'e de Haute-Alsace\\
Department of Computer Science, IRIMAS\\
F-68100 Mulhouse, France\\
$^{3}$Universit\'e de Strasbourg\\
\texttt{karim.hammoudi@uha.fr} \\
	   \And
Halim Benhabiles$^{4}$\\
  $^{4}$Univ. Lille, CNRS, Centrale Lille\\
  Univ. Polytechnique Hauts-de-France\\
 Yncrea Hauts-de-France \\
UMR 8520 - IEMN\\
Lille F-59000, France\\
  \texttt{halim.benhabiles@yncrea.fr} \\
	   \And
Mahmoud Melkemi$^{2,3}$\\
$^{2}$Universit\'e de Haute-Alsace\\
Department of Computer Science, IRIMAS\\
F-68100 Mulhouse, France\\
$^{3}$Universit\'e de Strasbourg\\
  \texttt{mahmoud.melkemi@uha.fr} \\
}

\begin{document}
\maketitle

\begin{abstract}
The wearing of the face masks appears as a solution for limiting the spread of COVID-19. 
In this context, efficient recognition systems are expected for checking that people faces are masked in regulated areas.
To perform this task, a large dataset of masked faces is necessary for training deep learning models towards detecting people wearing masks and those not wearing masks.
Some large datasets of masked faces are available in the literature. However, at the moment, there are no available large dataset of masked face images that permits to check if detected masked faces are correctly worn or not. Indeed, many people are not correctly wearing their masks due to bad practices, bad behaviors or vulnerability of individuals (e.g., children, old people). For these reasons, several mask wearing campaigns intend to sensitize people about this problem and good practices. In this sense, this work proposes three types of masked face detection dataset; namely, the Correctly Masked Face Dataset (CMFD), the Incorrectly Masked Face Dataset (IMFD) and their combination for the global masked face detection (MaskedFace-Net). Realistic masked face datasets are proposed with a twofold objective: i) to detect people having their faces masked or not masked, ii) to detect faces having their masks correctly worn or incorrectly worn (e.g.; at airport portals or in crowds). To the best of our knowledge, no large dataset of masked faces provides such a granularity of classification towards permitting mask wearing analysis. Moreover, this work globally presents the applied mask-to-face deformable model for permitting the generation of other masked face images, notably with specific masks. Our datasets of masked face images (137,016 images) are available at {\color{magenta}\href{https://github.com/cabani/MaskedFace-Net}{\textbf{https://github.com/cabani/MaskedFace-Net}}}.
\end{abstract}

\keywords{virus protection \and COVID-19 \and masked face dataset \and correct mask wearing \and masked face detection.}

\section{Introduction and motivation}

The wearing of the face masks appears as a solution for limiting the spread of COVID-19. 
In this context, efficient recognition systems are expected for checking that people faces are masked in regulated areas.
To perform this task, a large dataset of masked faces is necessary for training deep learning models towards detecting people wearing masks and those not wearing masks.
In this sense, some large datasets of face images with virus-related protection mask are available in the literature; e.g. the MAsked FAces dataset (MAFA) \cite{MAFA}, the Real-World Masked Face Dataset (RMFD\footnote{see ``Real-World Masked Face Dataset'' \url{https://github.com/X-zhangyang/Real-World-Masked-Face-Dataset}.}) and a masked face recognition dataset \cite{wang2020masked} composed of Masked Face Detection Dataset (MFDD), Real-world Masked Face Recognition Dataset (RMFRD) and Simulated Masked Face Recognition Dataset (SMFRD). 

\begin{figure*}[t!] \centering
\begin{subfigure}{.49\textwidth}
  \centering
  \includegraphics[width=8cm]{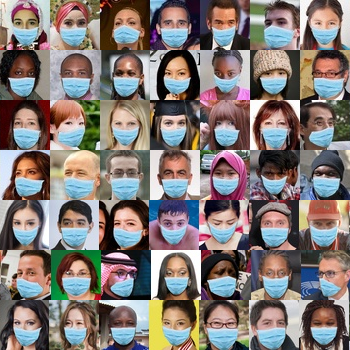}  
  \caption{Samples of correctly masked faces (dataset CMFD).}
  \label{mosaicok}
\end{subfigure}
\begin{subfigure}{.49\textwidth}
  \centering
  \includegraphics[width=8cm]{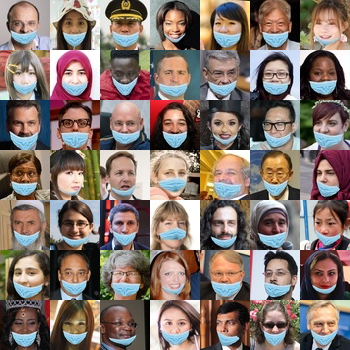}  
  \caption{Samples of incorrectly masked faces (dataset IMFD).}
  \label{mosaicko}
\end{subfigure}
\caption{A snapshot of the generated MaskedFace-Net image datasets: Fig.\ref{mosaicok} displays samples of correctly masked faces (dataset CMFD); Fig.\ref{mosaicko} displays face samples with a mix of incorrectly masked faces (dataset IMFD). For each set, 49 randomly sampled images are presented amongst approximately 70,000 generated images.}
\label{mosaic}
\end{figure*}

Besides, many people are not correctly wearing their masks due to bad practices, bad behaviors or vulnerability of individuals (e.g., children, old people). In this sense, several mask wearing campaigns intend to sensitize people about this problem and good practices \cite{AfricanUnion,infirmiere,Ivoire,lesoirBE}. In \cite{hammoudi}, a mobile application ``CheckYourMask'' has been designed towards permitting people to check if their mask is correctly worn or not by taking a selfie. The creation of a dataset with correctly/incorrectly worn mask classes has been suggested. In \cite{makeml}, a dataset composed of images with individual or multiple masked faces (853 images) has been proposed towards creating detection model taking into account the improperly masked faces. In the present case, we propose a relatively large dataset of 137,016 masked face images that is divided into two masked face categories; correctly worn and incorrectly worn (see samples in Fig.\ref{mosaicok} and Fig.\ref{mosaicko}, respectively).   

Specifically, this work proposes three types of masked face detection dataset; namely, the Correctly Masked Face Dataset (CMFD), the Incorrectly Masked Face Dataset (IMFD) and their combination (MaskedFace-Net) for the masked face detection (see dataset structure in Fig.\ref{algo1}). Realistic masked face datasets are proposed with a twofold objective: i) to detect people having their faces masked or not masked, ii) to detect faces having their masked correctly worn or incorrectly worn (e.g.; at airport portals or in crowds). To best of our knowledge, no large dataset of masked faces provides such a granularity of classification towards permitting mask wearing analysis. Moreover, this work globally presents the applied mask-to-face deformable model for permitting the generation of other masked face images, notably with specific masks.


\section{Applied mask-to-face deformable model}

The dataset of face images Flickr-Faces-HQ\footnote{see ``dataset of face images Flickr-Faces-HQ (FFHQ)'' \url{https://github.com/NVlabs/ffhq-dataset}.} (FFHQ) has been selected as a base for creating an enhanced dataset MaskedFace-Net composed of correctly and incorrectly masked face images. Indeed, FFHQ contains 70000 high-quality images of human faces in PNG images at $1024\times1024$ resolution and is free of use. The FFHQ dataset offers a lot of variety in terms of age, ethnicity, viewpoint, lighting, and image background. It was originally created as a benchmark for generative adversarial networks (GAN) \cite{karras2018stylebased}. 

\begin{figure}[t] \centering
\begin{subfigure}{.3\textwidth}
  \centering
  \includegraphics[height=4cm]{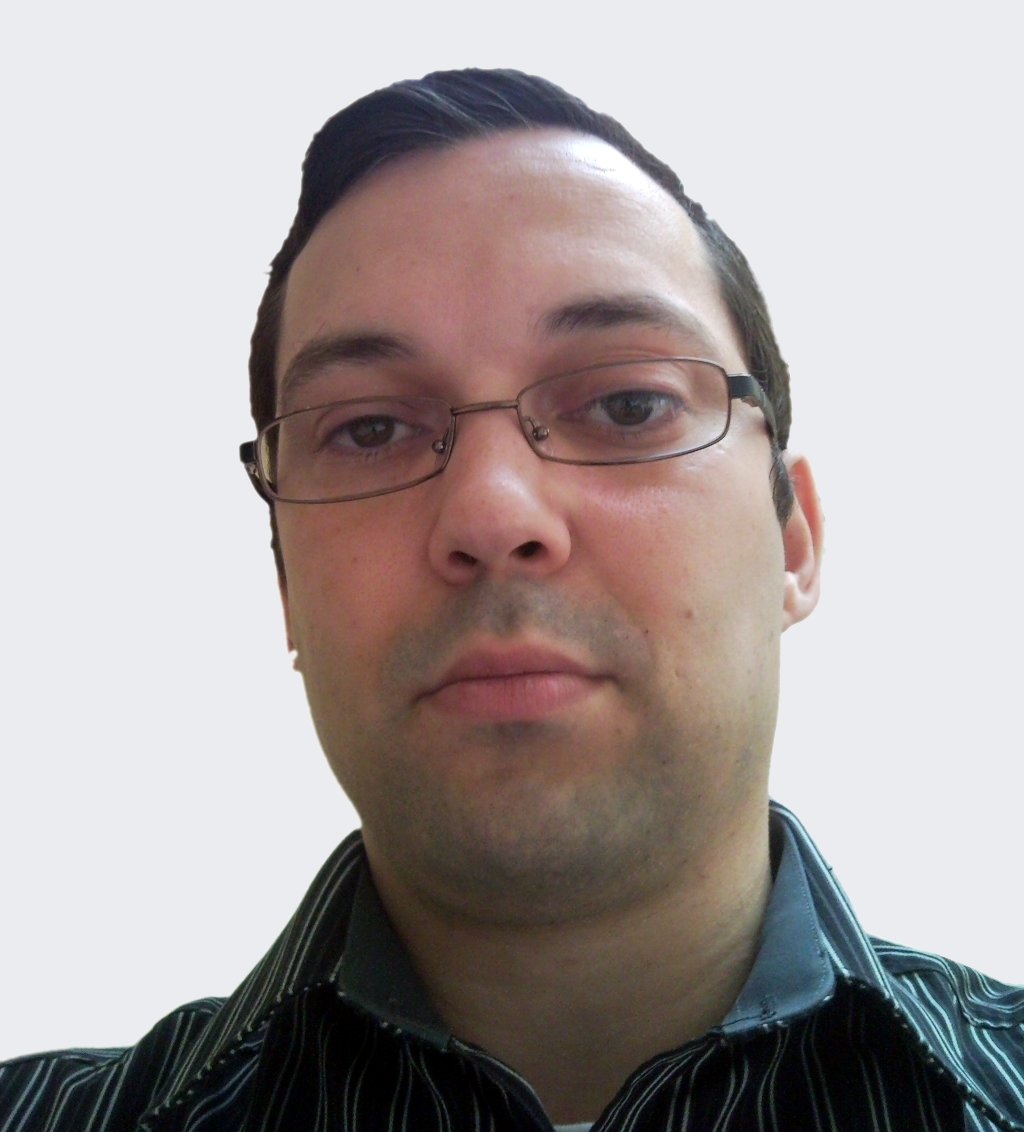}  
  \caption{Face image (first author).}
  \label{1}
\end{subfigure}
\begin{subfigure}{.3\textwidth}
  \centering
  \includegraphics[height=4cm]{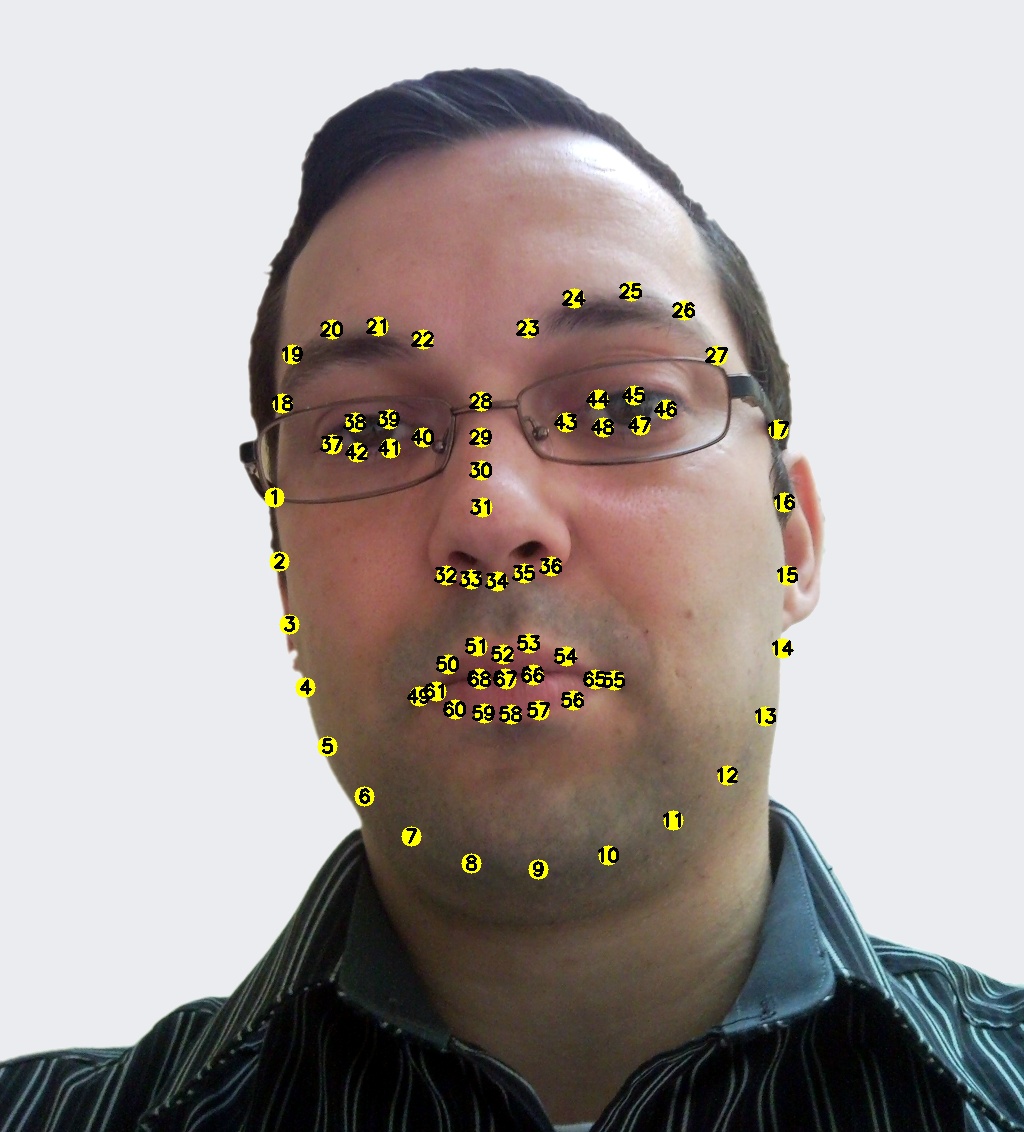}  
  \caption{Detected landmarks.}
  \label{2}
\end{subfigure}
\begin{subfigure}{.3\textwidth}
  \centering
  \includegraphics[height=4cm]{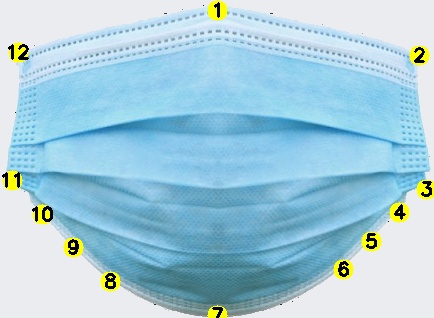}  
  \caption{Annotated landmarks.}
  \label{3}
\end{subfigure}
\caption{Sample of face image, reference mask image and considered landmarks.}
\label{landmarks}
\end{figure}

Our MaskedFace-Net dataset has been created by defining a mask-to-face deformable model. A pseudo-code of the global principle for generating MaskedFace-Net is shown in Fig.\ref{algo2} with respect to outputs depicted in Fig.\ref{algo1}. For each face image of FFHQ (e.g. Fig. \ref{1}), Haar feature-based cascade classifiers are applied for detecting a region of interest (detection of face rectangle). Then, a specific key point detector ``shape predictor 68 face landmarks\footnote{see ``Facial point annotations'' \url{https://ibug.doc.ic.ac.uk/resources/facial-point-annotations/}.}\footnote{see ``shape predictor 68 face landmarks.dat.bz2'' \url{https://github.com/davisking/dlib-models\#shape_predictor_68_face_landmarksdatbz2}.}'' (model derived from \cite{Sagonas}) is applied to the detected region of interest and permits to automatically detect 68 landmarks of the facial structure (see sample in Fig. \ref{2}). Besides, an image of a conventional face protection mask (single-use blue face protection mask) has been selected as a reference image for the mapping (see sample in Fig. \ref{3}). For this latter, 12 key points have manually been annotated for delineating the mask area (polygonal area).

\begin{figure*}[t!] \centering
\begin{subfigure}{.49\textwidth}
  \centering
  \begin{forest}
for tree={circle,draw, l sep=20pt}
[MaskedFace-Net,blue 
    [CMFD,teal, edge label={node[midway,left] {Correctly masked}}
    ]
    [IMFD,teal, edge label={node[midway,right] {Incorrectly masked}}
      [IMFD1, orange,edge label={node[midway,left] {Uncovered chin}} ]  
      [IMFD2,orange,label=below:Uncovered nose]
      [IMFD3,orange,edge label={node[midway,right] {Uncovered nose and mouth}} ] 
  ] 
]
\end{forest}
  \caption{View of the MaskedFace-Net data tree.}
    \label{algo1}
\end{subfigure} 
\scalebox{0.91}{\begin{subfigure}{.49\textwidth}
  \centering
		\footnotesize
    \begin{algorithmic}
        \Function{generating MaskedFace-Net}{$FFHQ,mask$}
            \For{$(each~face~image \in FFHQ)$} 
						\State \textbf{FaceDetection()}
						\State \textbf{FacialLandmarksDetection()}
								\If{$(CMFD~targeted)$} 
									\State \textbf{Match}($landmards$)
									\State \textbf{Map}($mask,face$)
								\EndIf
								\If{$(IMFD1~targeted)$}
									\State \textbf{Match}($landmards$)
									\State \textbf{Map}($mask,face$)
								\EndIf
								\If{$(IMFD2~targeted)$} 
									\State \textbf{Match}($landmards$)
									\State \textbf{Map}($mask,face$)
								\EndIf
								\If{$(IMFD3~targeted)$} 
									\State \textbf{Match}($landmards$)
									\State \textbf{Map}($mask,face$)
								\EndIf
            \EndFor
						\State \Return MaskedFace-Net
        \EndFunction
    \end{algorithmic}
		  \caption{{\normalsize Pseudo-code for the generation of MaskedFace-Net.}}
    \label{algo2}
\end{subfigure}}
\caption{Fig.\ref{algo1} depicts the structure of the generated MaskedFace-Net dataset. Fig.\ref{algo2} shows a pseudo-code of the mask-to-face deformable model applied for generating outputs Fig.\ref{algo1} of the MaskedFace-Net dataset.}
\label{mosaic}
\end{figure*}

At this stage, four types of mask-to-face mapping have been defined with respect to targeted cases (see Fig.\ref{algo1}), namely mask covering the nose, mouth and chin (i.e. mask correctly worn), mask only covering the nose and mouth; mask only covering mouth and chin and mask under the mouth (i.e. three cases of mask incorrectly worn). For each type of mask-to-face mapping (CMFD, IMFD1, IMFD2 or IMFD3), a subset of 12 facial key points is retained from the 68 landmarks automatically detected; then matched to the 12 mask key points. By this way, the mask can fit specific areas of the face for each targeted case. Hence, a mask-to-face deformable mask model has been created to generate MaskedFace-Net. Moreover, each targeted case can have up to 2 key points of the mask (amongst 12 key points) that have their locations randomly displaced in a limited perimeter. In particular, this tolerance allows to act on the height of the mask on the face and then to bring more realism to the generated dataset.

\begin{figure}[t] \centering
\begin{subfigure}{.2\textwidth}
  \centering
  \includegraphics[width=3cm]{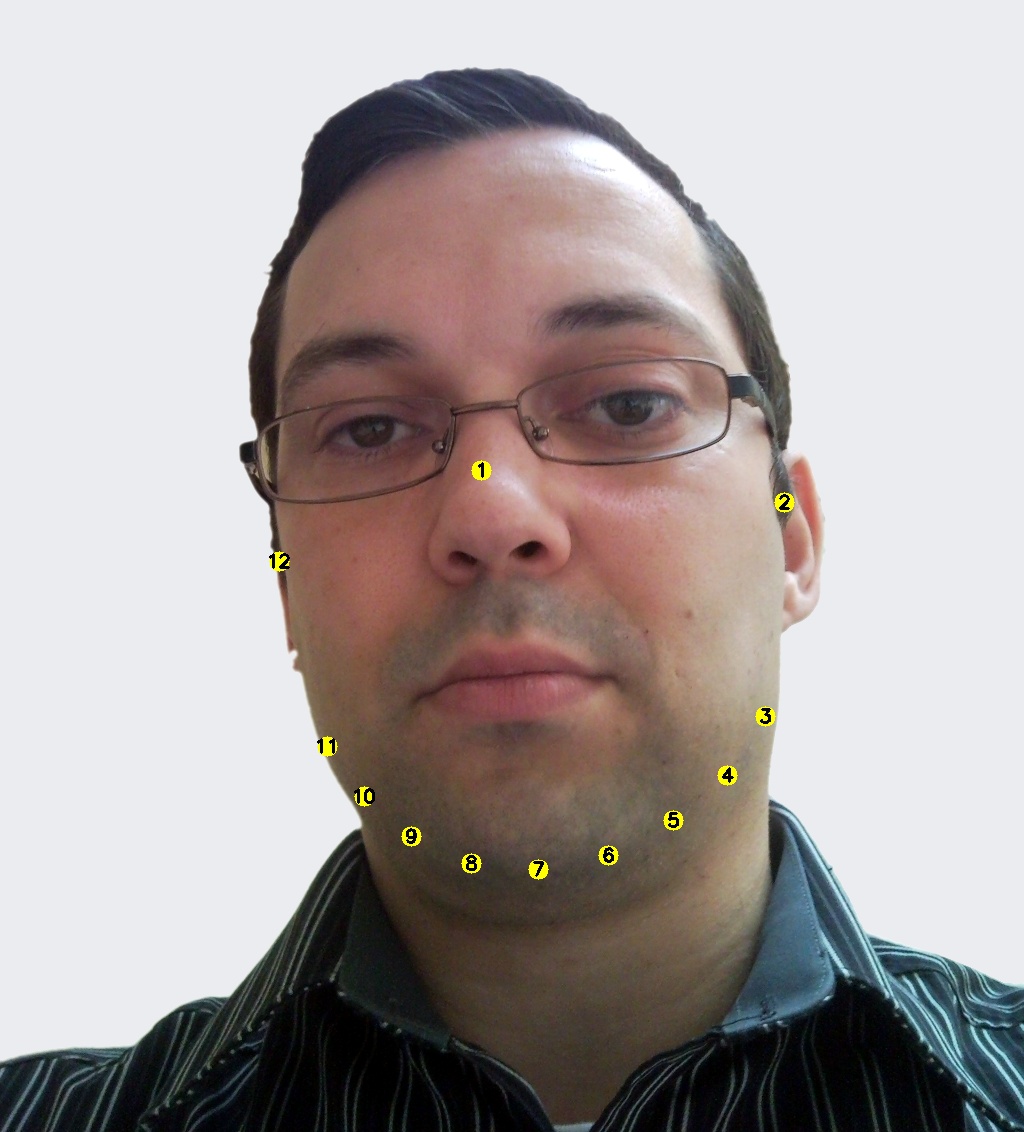}  
  \caption{Detected landmarks.}
  \label{detect1}
\end{subfigure}
\begin{subfigure}{.2\textwidth}
  \centering
  \includegraphics[width=3cm]{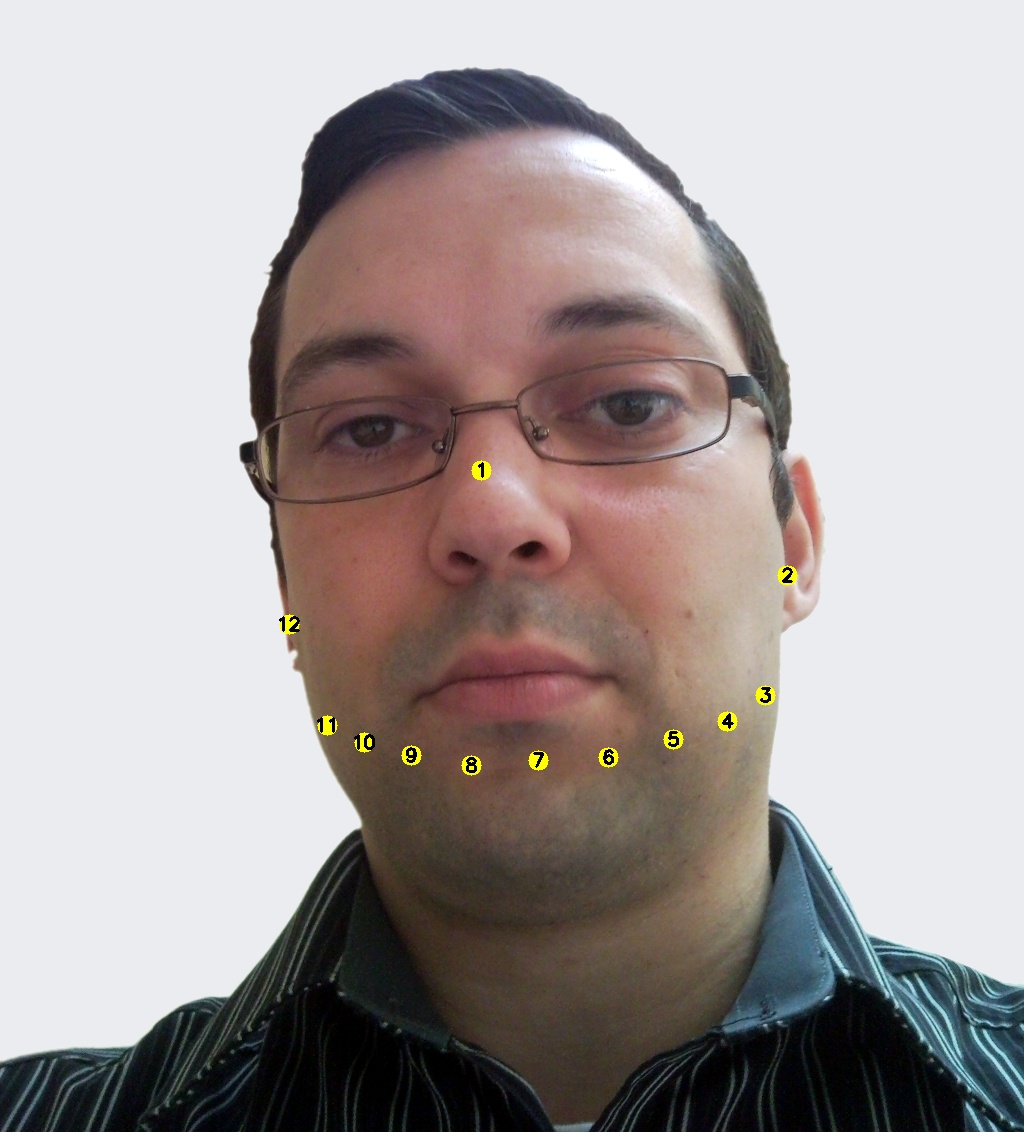}  
  \caption{Detected landmarks.}
  \label{detect2}
\end{subfigure}
\begin{subfigure}{.2\textwidth}
  \centering
  \includegraphics[width=3cm]{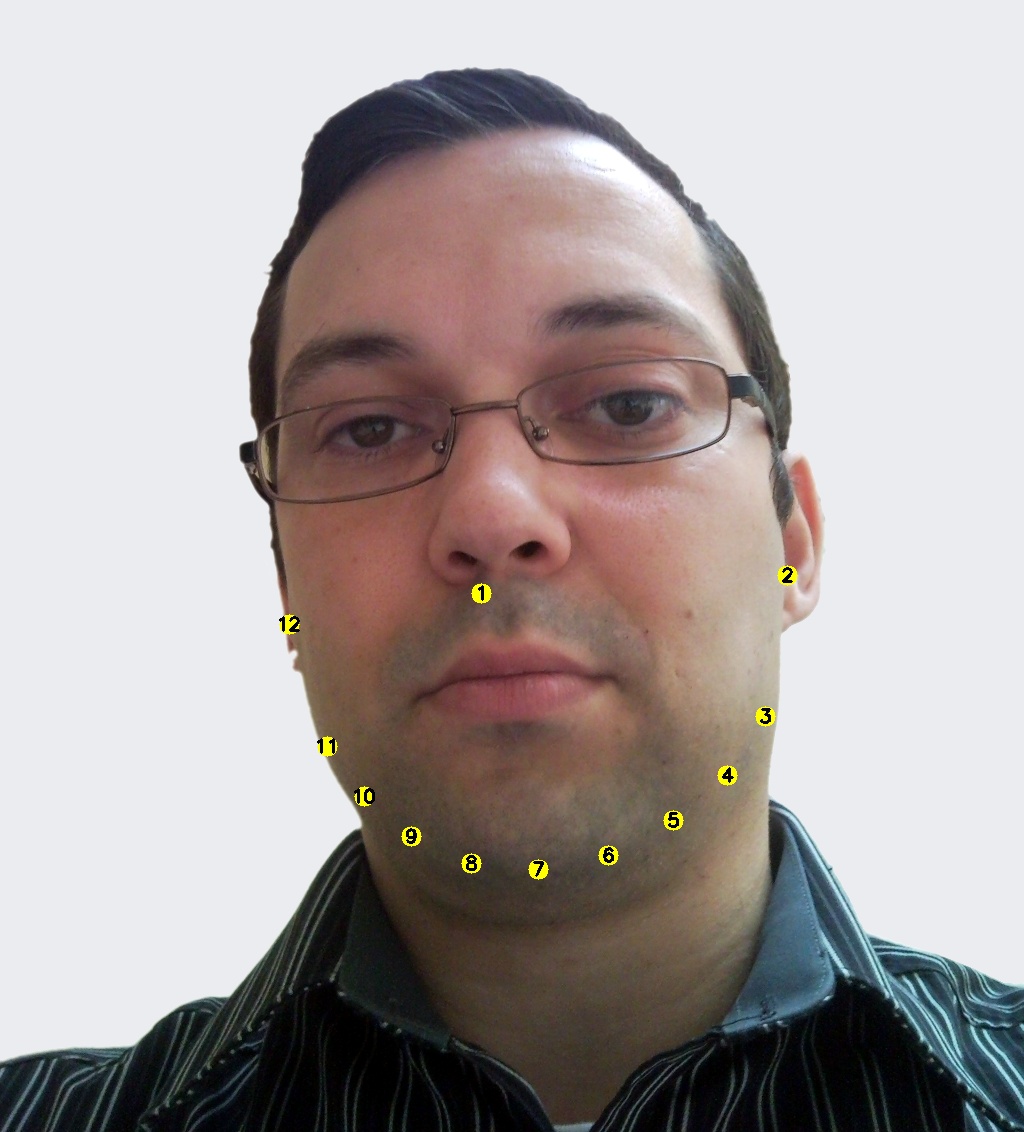}  
  \caption{Detected landmarks.}
  \label{detect3}
\end{subfigure}
\begin{subfigure}{.2\textwidth}
  \centering
  \includegraphics[width=3cm]{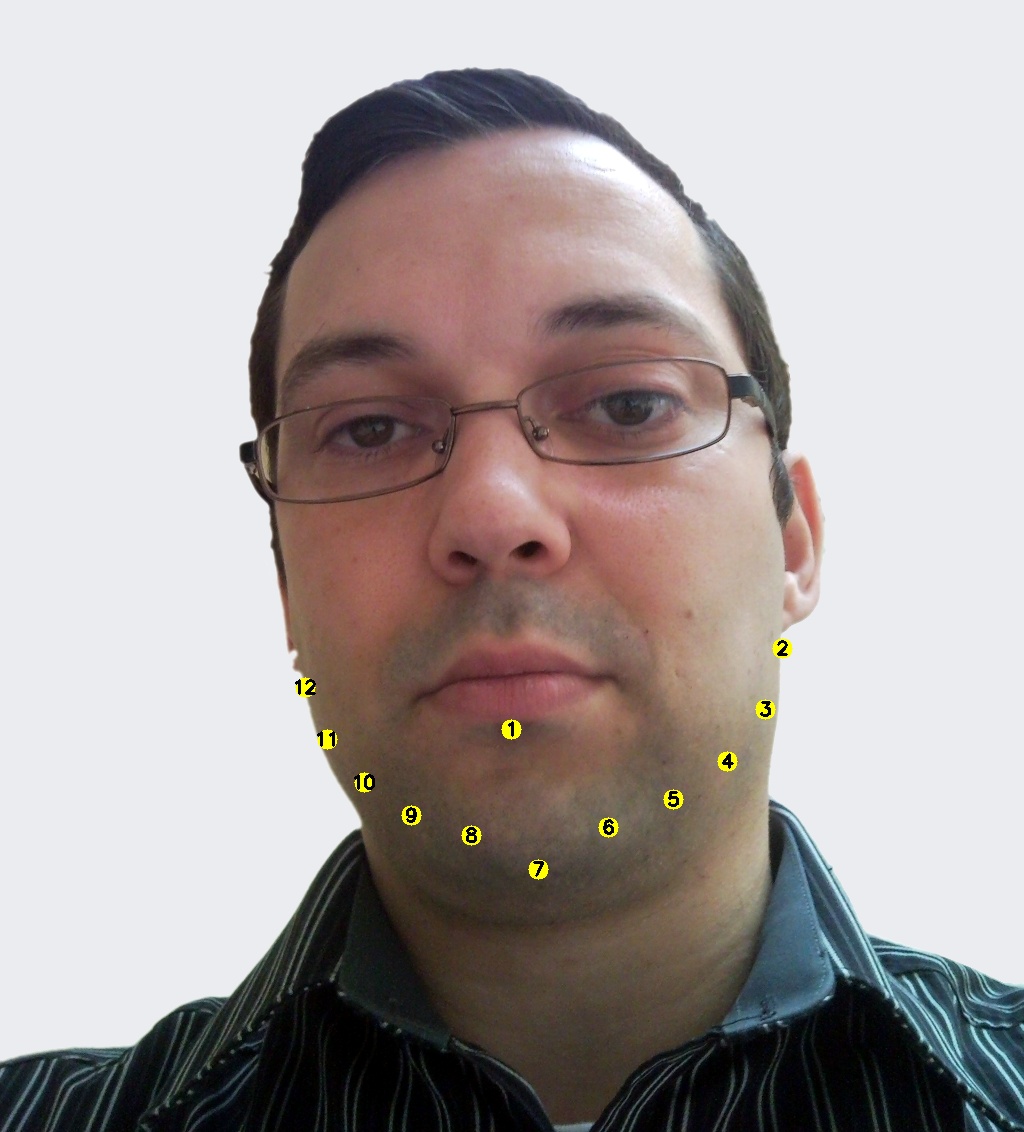}  
  \caption{Detected landmarks.}
  \label{detect4}
\end{subfigure}
\begin{subfigure}{.2\textwidth}
  \centering
  \includegraphics[width=3cm]{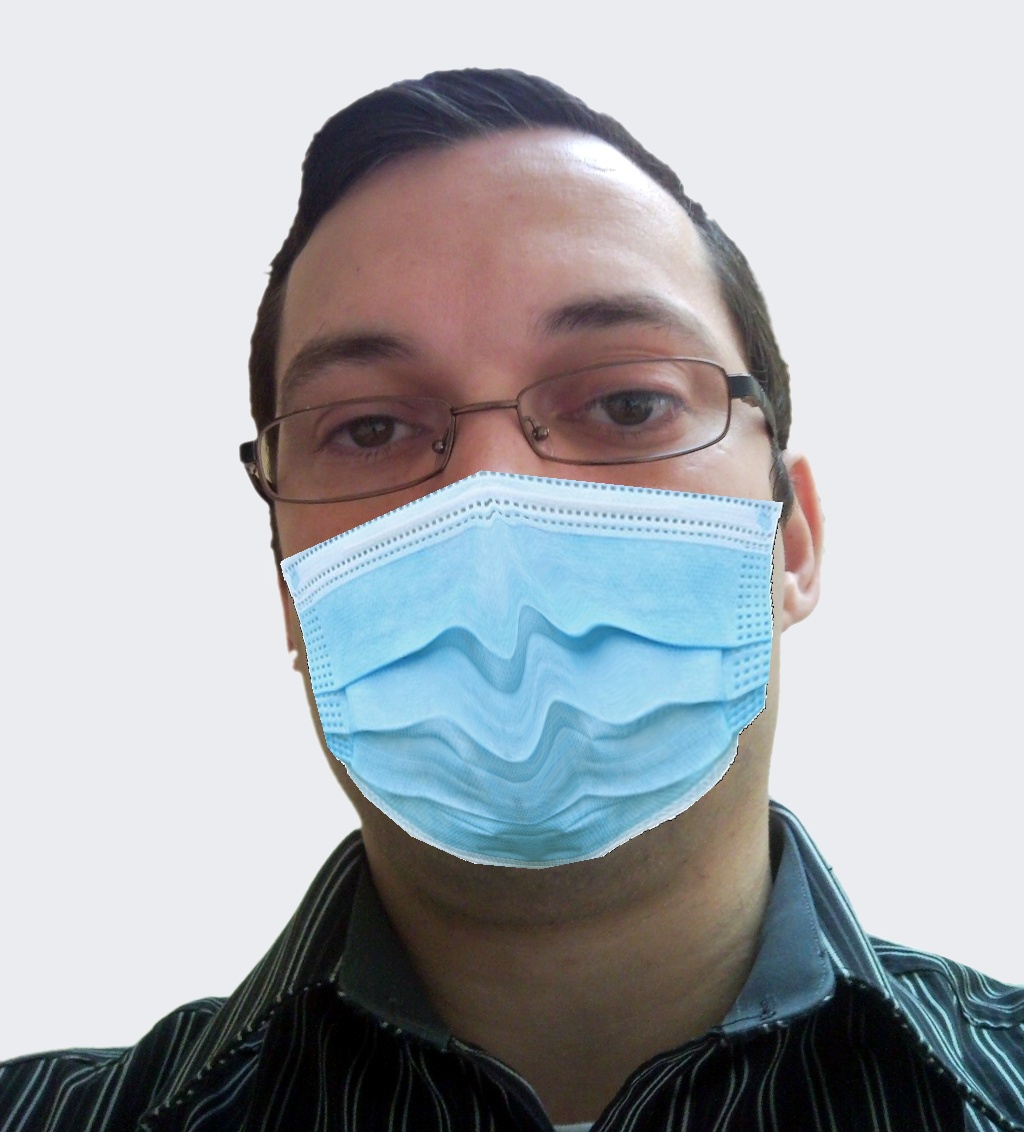}  
  \caption{Correctly masked.}
  \label{project1}
\end{subfigure}
\begin{subfigure}{.2\textwidth}
  \centering
  \includegraphics[width=3cm]{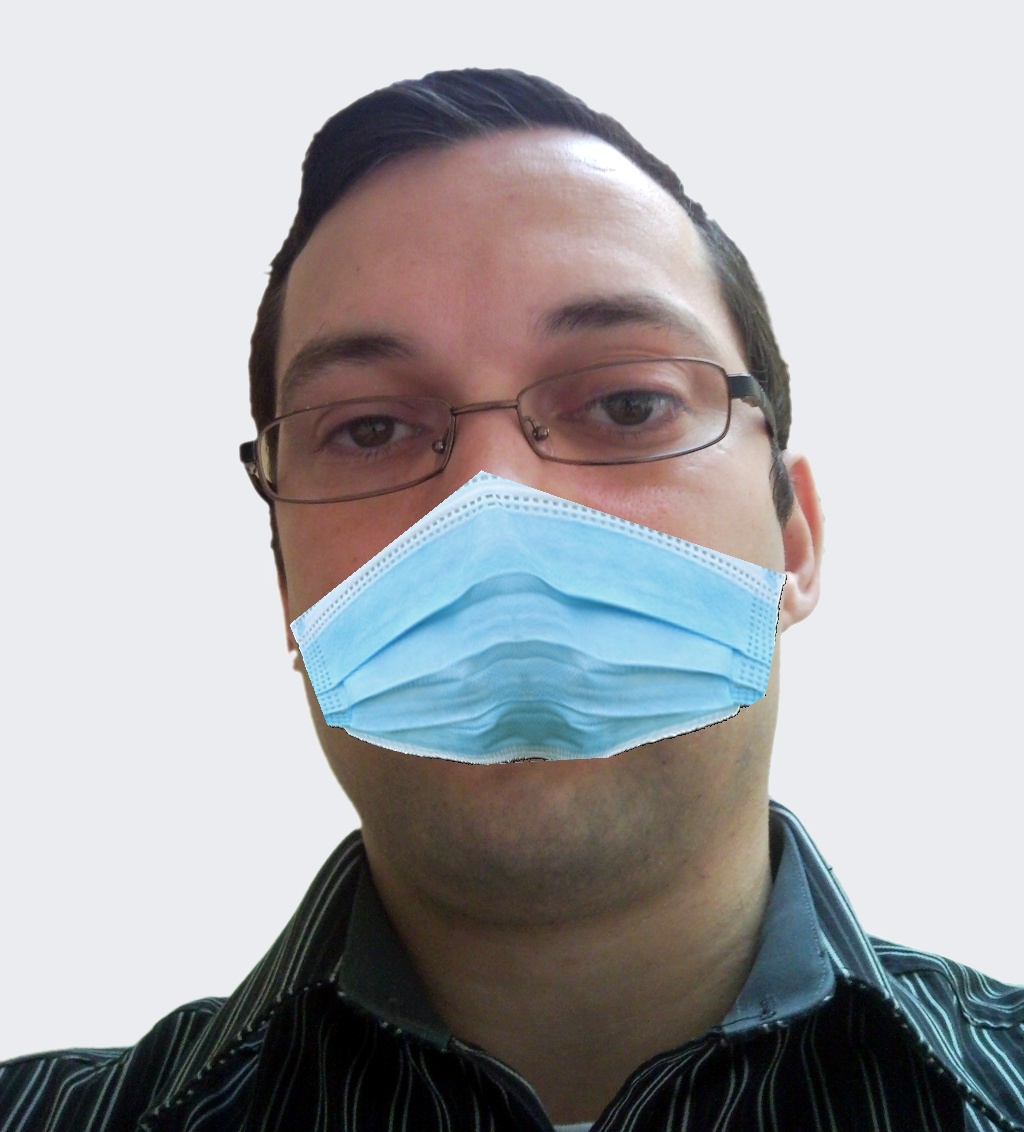}  
  \caption{Incorrectly masked.}
  \label{project2}
\end{subfigure}
\begin{subfigure}{.2\textwidth}
  \centering
  \includegraphics[width=3cm]{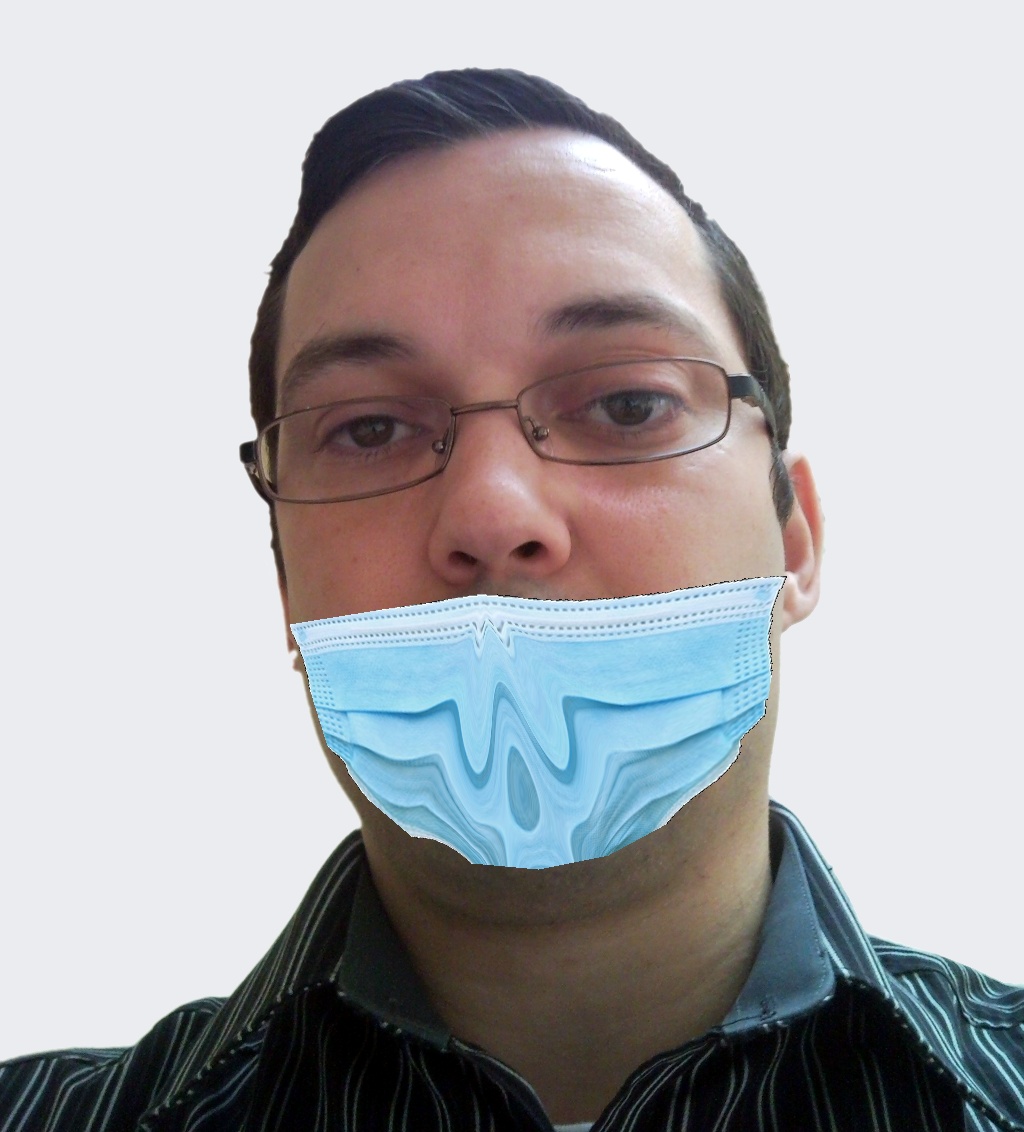}  
  \caption{Incorrectly masked.}
  \label{project3}
\end{subfigure}
\begin{subfigure}{.2\textwidth}
  \centering
  \includegraphics[width=3cm]{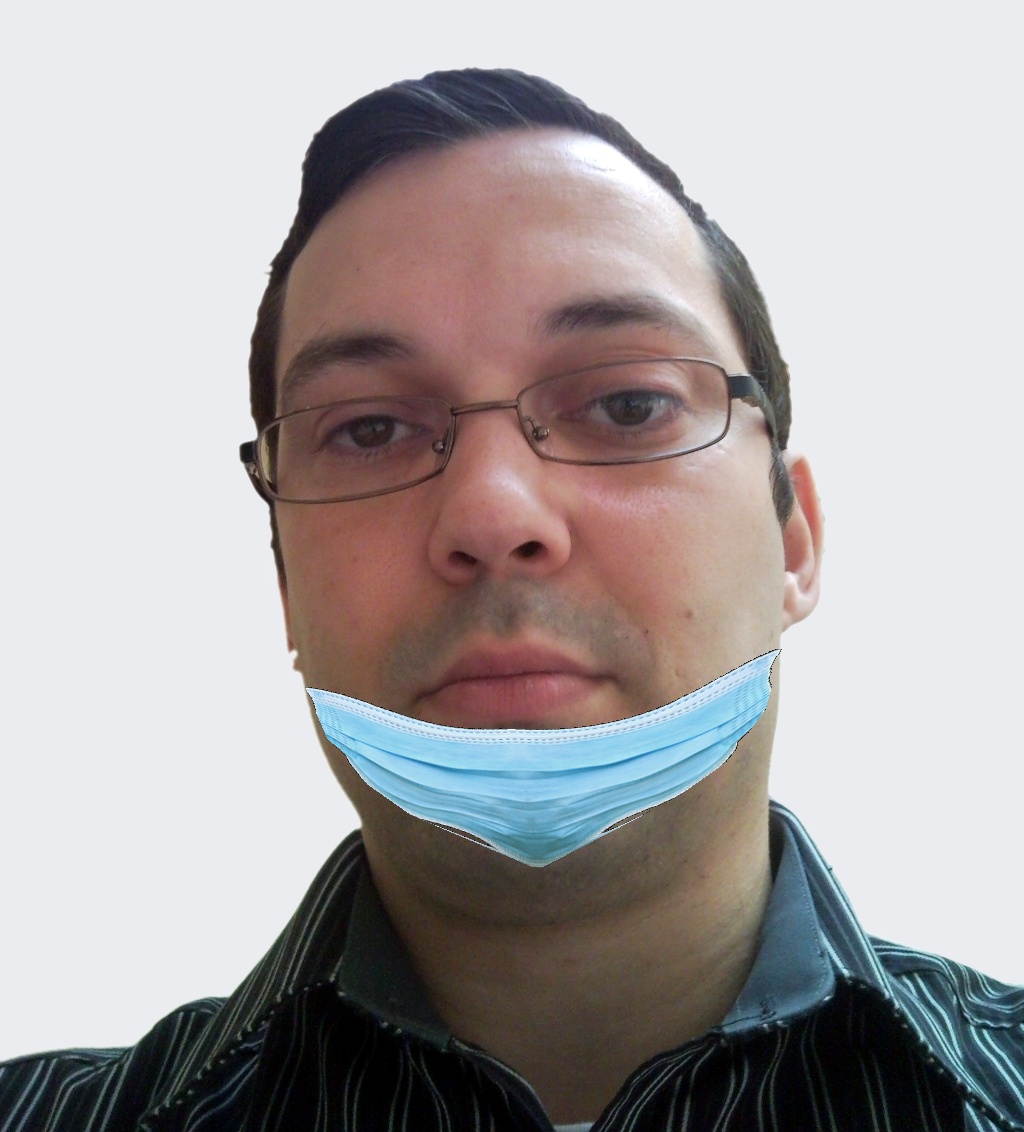}  
  \caption{Incorrecly masked.}
  \label{project4}
\end{subfigure}
\caption{(row $1$) Results of landmarks detection with respect to the defined types of mask-to-face mapping. (row $2$) Corresponding results of mask-to-face mapping driven by the proposed deformable model towards generating the MaskedFace-Net dataset.}
\label{landmarks}
\end{figure}

Finally, a homography transformation which relies on the defined point-to-point correspondence of landmarks between mask image and face image is applied for mapping mask pixels over the targeted facial areas. Instances of produced face landmarks and corresponding mask-to-face mapping are displayed for each type in Fig.\ref{detect1} and Fig.\ref{detect2}, Fig.\ref{detect3}, Fig.\ref{detect4} and Fig.\ref{project1}, Fig.\ref{project2}, Fig.\ref{project3}, Fig.\ref{project4}, respectively.  

For information, Fig.\ref{algo2} illustrates a nominal scenario of face-related detection. However, some faces of FFHQ have not been processed (177 images) since face occlusions (e.g., arms, hands) made the face detection failing  (i.e. no detected face rectangle). After the face detection, the MaskedFace-Net dataset contained 139,646 images. Moreover, a manual filtering has been operated for deleting detected face images having their mask incorrectly mapped in reason of failing landmark detection. Indeed, erroneous landmark detection occurs when the visibility of the facial contours is limited (e.g. for profile views of detected faces). Hence, the proposed MaskedFace-Net dataset contains 137,016 masked face images. The MaskedFace-Net dataset is composed of 49\% of correctly masked faces (67,193 images) and 51\% of incorrectly masked faces (69,823 images). For this latter set, approximately 80\% represents faces with only mouth and chin masked, 10\% with only nose and mouth masked and 10\% with only chin masked.
 
We emphasize that a raw mask-to-face mapping has been applied to the FFHQ dataset. In particular, no images have been filtered according to specific parameters (e.g. age). However, the file naming of MaskedFace-Net includes the one given by the FFHQ dataset. Hence, correspondence in between the FFHQ and MaskedFace-Net can be established towards related filtering. 

It is worth mentioning that the minimum age for mask wearing depends on applicable laws in concerned countries. For instance, the mask wearing is compulsory from 6 years old in Spain, 11 years old in France, 12 years old in Belgium under certain conditions \cite{media}. Between 2 and 11 years old, opinions differ \cite{media1}. Since FFHQ contains face images of all ages, it is also the case for masked face image of MaskedFace-Net. Such datasets could then be exploited for detecting children in crowds that wear a mask under the recommended limit of age. 
 
\section{Conclusion}

A large dataset of 137,016 quality masked face images has been produced and made available online. MaskedFace-Net can be seen as a benchmark dataset for creating machine learning models related to the mask wearing analysis; notably, detecting the presence of mask or not over detected face images, the correct or incorrect wearing for detected masked faces. MaskedFace-Net can then be used for enhancing vision-based monitoring systems towards several applications such as checking the respect of laws related to the mask wearing or generating crowd statistics. Moreover, the method used for the generation of MaskedFace-Net has been described for permitting the generation of masked face images by using other types of mask.

\section{Disclaimer}

In no case the contributors of this work could be held responsible for any incident when using the MaskedFace-Net dataset or masks.

\bibliographystyle{IEEEtran} 

\bibliography{references}  

\begin{thebibliography}{10}
\providecommand{\url}[1]{#1}
\csname url@samestyle\endcsname
\providecommand{\newblock}{\relax}
\providecommand{\bibinfo}[2]{#2}
\providecommand{\BIBentrySTDinterwordspacing}{\spaceskip=0pt\relax}
\providecommand{\BIBentryALTinterwordstretchfactor}{4}
\providecommand{\BIBentryALTinterwordspacing}{\spaceskip=\fontdimen2\font plus
\BIBentryALTinterwordstretchfactor\fontdimen3\font minus
  \fontdimen4\font\relax}
\providecommand{\BIBforeignlanguage}[2]{{%
\expandafter\ifx\csname l@#1\endcsname\relax
\typeout{** WARNING: IEEEtran.bst: No hyphenation pattern has been}%
\typeout{** loaded for the language `#1'. Using the pattern for}%
\typeout{** the default language instead.}%
\else
\language=\csname l@#1\endcsname
\fi
#2}}
\providecommand{\BIBdecl}{\relax}
\BIBdecl

\bibitem{MAFA}
S.~{Ge}, J.~{Li}, Q.~{Ye}, and Z.~{Luo}, ``Detecting masked faces in the wild
  with lle-cnns,'' in \emph{2017 IEEE Conference on Computer Vision and Pattern
  Recognition (CVPR)}, 2017, pp. 426--434.

\bibitem{wang2020masked}
Z.~Wang, G.~Wang, B.~Huang, Z.~Xiong, Q.~Hong, H.~Wu, P.~Yi, K.~Jiang, N.~Wang,
  Y.~Pei, H.~Chen, Y.~Miao, Z.~Huang, and J.~Liang, ``Masked face recognition
  dataset and application,'' 2020, arXiv:2003.09093.

\bibitem{AfricanUnion}
\BIBentryALTinterwordspacing
{Africa Centres for Disease Control and Prevention - Africa {CDC}, African
  Union}, ``How to wear a face mask correctly,'' 2020.
  \url{https://africacdc.org/download/how-to-wear-a-face-mask-correctly/}
\BIBentrySTDinterwordspacing

\bibitem{infirmiere}
\BIBentryALTinterwordspacing
J.~Bouteiller, ``Coronavirus. comment bien porter son masque ? les conseils
  d'une infirmi\`ere de la m\'etropole de {L}ille,'' 2020.
  \url{https://actu.fr/hauts-de-france/lille_59350/coronavirus-comment-bien-porter-masque-conseils-dune-infirmiere_32651335.html}
\BIBentrySTDinterwordspacing

\bibitem{Ivoire}
\BIBentryALTinterwordspacing
{Action Sant\'e-Social C\^ote d'Ivoire}, ``Comment bien mettre son masque,''
  2020.
  \url{https://www.facebook.com/110412877115436/photos/comment-bien-mettre-son-masque/154573562699367/}
\BIBentrySTDinterwordspacing

\bibitem{lesoirBE}
\BIBentryALTinterwordspacing
L.~Colart, ``Le port du masque: les gestes \`a faire et ne pas faire,'' 2020,
  \url{https://www.lesoir.be/sites/default/files/dpistyles_v2/ena_16_9_in_line/2020/04/21/node_296003/27512244/public/2020/04/21/B9723268640Z.1_20200421182927_000+GBLFTKA92.1-0.jpg?itok=vge-65yl1587734455}.
   \url{https://plus.lesoir.be/296003/article/2020-04-21/le-port-du-masque-les-gestes-faire-et-ne-pas-faire}
\BIBentrySTDinterwordspacing

\bibitem{hammoudi}
K.~Hammoudi, A.~Cabani, H.~Benhabiles, and M.~Melkemi, ``Validating the correct
  wearing of protection mask by taking a selfie: design of a mobile application
  ``{C}heck{Y}our{M}ask'' to limit the spread of {COVID}-19,'' \emph{Computer
  Modeling in Engineering \& Sciences}, 2020.

\bibitem{makeml}
\BIBentryALTinterwordspacing
``Mask dataset.''  \url{https://makeml.app/datasets/mask}
\BIBentrySTDinterwordspacing

\bibitem{karras2018stylebased}
T.~Karras, S.~Laine, and T.~Aila, ``A style-based generator architecture for
  generative adversarial networks,'' 2018, arXiv:1812.04948.

\bibitem{Sagonas}
C.~Sagonas, E.~Antonakos, G.~Tzimiropoulos, S.~Zafeiriou, and M.~Pantic,
  ``\BIBforeignlanguage{Undefined}{300 faces in-the-wild challenge: database
  and results},'' \emph{\BIBforeignlanguage{Undefined}{Image and vision
  computing}}, vol.~47, pp. 3--18, 2016.

\bibitem{media}
\BIBentryALTinterwordspacing
{RTBF}, ``Le masque obligatoire d{\`{e}}s 6 ans en espagne, 11 ans en france,
  12 ans en belgique : pourquoi tant de diff{\'{e}}rences ?'' 2020.
  \url{https://www.rtbf.be/info/societe/detail_le-masque-obligatoire-des-6-ans-en-espagne-11-ans-en-france-12-ans-en-belgique-pourquoi-tant-de-differences?id=10549655}
\BIBentrySTDinterwordspacing

\bibitem{media1}
\BIBentryALTinterwordspacing
C.~Daclin, ``Coronavirus : o{\`{u}} et {\`{a}} quel {\^{a}}ge les enfants
  doivent-ils porter le masque ?'' 2020.
  \url{https://www.rtl.fr/actu/bien-etre/coronavirus-ou-et-a-quel-age-les-enfants-doivent-ils-porter-le-masque-7800688133}
\BIBentrySTDinterwordspacing

\end{thebibliography}

%
%
%
%
%

\end{document}